# A Two-Stage Dual-Path Framework for Text Tampering Detection and Recognition

Guandong Li[1*], Xian Yang[1], Wenpin Ma[1]

1.*Suning, Xuanwu, Nanjing, 210042, Jiangsu, China.

*Corresponding author(s). E-mail(s): leeguandon@gmail.com

*Abstract*: Document tamper detection has always been an important aspect of tamper detection. Before the advent of deep learning, document tamper detection was difficult. We have made some explorations in the field of text tamper detection based on deep learning. Our Ps tamper detection method includes three steps: feature assistance, audit point positioning, and tamper recognition. It involves hierarchical filtering and graded output (tampered/suspected tampered/untampered). By combining artificial tamper data features, we simulate and augment data samples in various scenarios (cropping with noise addition/replacement, single character/space replacement, smearing/splicing, brightness/contrast adjustment, etc.). The auxiliary features include exif/binary stream keyword retrieval/noise, which are used for branch detection based on the results. Audit point positioning uses detection frameworks and controls thresholds for high and low density detection. Tamper recognition employs a dual-path dual-stream recognition network, with RGB and ELA stream feature extraction. After dimensionality reduction through self-correlation percentile pooling, the fused output is processed through vlad, yielding an accuracy of 0.804, recall of 0.659, and precision of 0.913.

1.Introduction

Digital images are now easily manipulated and often in visually imperceptible ways. Image tampering has negative impacts on many aspects of modern society, such as fake news, online rumors, insurance fraud, etc. [1]. Copy-move (copying and moving an element from one area of a given image to another) [2], splicing (copying and pasting an element from one image onto another) [3], and restoration (removing unwanted elements) [4] are three common types of image manipulation that can lead to misunderstandings of visual content [5,6]. Our goal is to distinguish whether a certain region in an authentic image has been manipulated.

The document tampering we are focusing on in this article is a more challenging branch of image tamper detection. Image tamper detection is more common and widespread in natural scene images. Document tampering is often hidden in areas such as digital amounts, names, and times. The tampered areas are very small, the types of images vary greatly, extreme data imbalances exist, and the human eye cannot discern whether an image has been tampered with, nor the corresponding area. Furthermore, auditors also need auxiliary tools to help identify tampering, and there are various unknown adversarial attacks. Currently, we rarely see papers specifically targeting document tamper detection. Therefore, in order to address these issues in practical industry, we have proposed a multi-stage cascaded framework for text tamper detection, dividing the Ps tamper detection into three steps, including feature assistance, audit point positioning, tamper recognition, hierarchical filtering, and graded output. Additionally, the scarcity of data for document tamper detection poses a significant challenge. By combining artificial tamper data features, we simulate and augment data samples in various scenarios (cropping with noise addition/replacement, single character/space replacement, smearing/splicing, brightness/contrast adjustment, etc.)

In actual business scenarios, we have designed multiple sets of strategies at the feature-assisted layer to pre-screen the input for global feature selection through strategy combinations, which is often very useful. The strategy combination often takes into account traditional tamper detection methods, such as using EXIF to classify input images.The next step is audit point positioning. In the business scenario of document tamper detection, only sensitive positions are often audited, such as amounts and names. The positioning of audit points is not only a business requirement but also important in the algorithm's process design. Document tampering seeks differences between text and text, and between text and background. By focusing on these differences, we can find feature clues. In a complete image, weak features are often ignored by downsampling, and differential features in small feature maps often do not receive enough attention. Therefore, it is important to focus on strengthening the attention to local areas. In order to not disrupt the features after audit point positioning, we have designed a cropping strategy to non-destructively crop the regions for tamper recognition. In the detection network for audit point positioning, we use different thresholds to detect the outputs at different levels of the feature-assisted layer. We conduct high-density detection for images that are highly likely to be tampered with, and low-density detection for images with a lower likelihood of tampering. This is a detection strategy. Finally, tamper recognition is the most crucial part of the entire process. ManTraNet [7] discusses different forms of tampering types and network designs from various perspectives, including JPEG compression, edge inconsistencies, noise patterns, color consistency, visual similarity, EXIF consistency, and camera model. It is a common practice to use prior features to enhance tampering clues. In RGBN [8], SRM [9] is also used for extracting noise flow features. In our tamper recognition system, we use the ELA branch to form a dual-path recognition network, with RGB and ELA flow feature extraction. After dimensionality reduction through self-correlation percpooling [10], fusion, and output through NetVLAD [11], the network remains designed as a semantic segmentation network.

Our contributions are as follows:
1.We proposed a multi-stage cascaded framework for tamper detection in document scenarios.
2.We introduced a dual-path network based on document tamper detection.

2.Related works

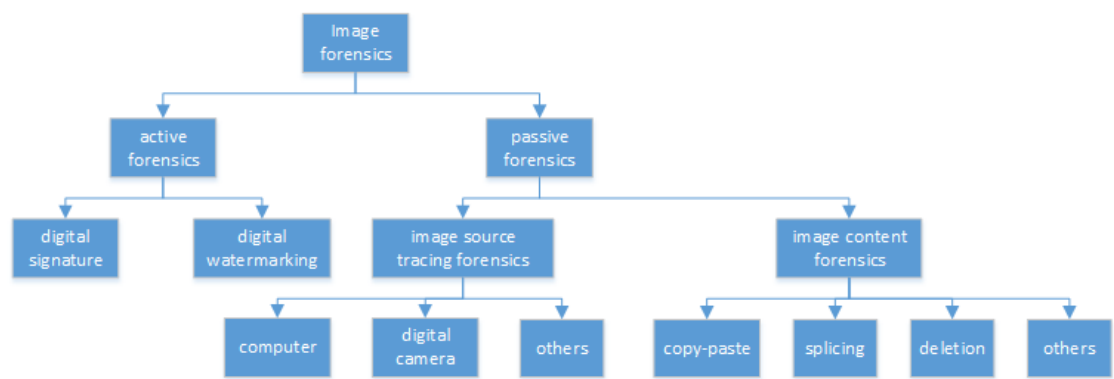

Fig1： Diagram of the domain division of digital forensics.

In terms of technical categories, tamper detection belongs to the field of image forensics, while document tamper detection falls under image content forensics in passive forensics. Passive image forensics focuses on subtle changes in tamper boundaries, judging the authenticity of an image based on changes in boundary artifacts and statistical features.

2.1 Traditional methods

Untampered authentic images typically exhibit continuity and integrity in their intrinsic statistical features, while these continuity features are disrupted in tampered images. Therefore, by analyzing the differences between the original and tampered images in these statistical intrinsic features (such as pixel mean values, RGB correlations, wavelet domain statistical features, etc.), one can determine the authenticity of an image. This approach usually involves manually designing certain statistical features, quantitatively describing the extracted features, using machine learning methods (such as SVM) for classification training, and finally making decisions based on the trained classifier. Below, we will introduce some traditional tamper detection methods.

Traditional tamper detection types include image splicing, which refers to merging a portion of the donor image into the source image to create a new tampered image. Copy-move detection involves operations within a single image, where real and tampered regions are very similar in statistical properties, making it challenging to utilize inherent device properties and most image statistical features. There are generally two main methods: based on region boundary artifacts and based on region similarity. To better suppress image semantic information, edge features are extracted, SRM high-pass filters are used to extract high-frequency information from the image, and then a sliding window method is employed to input the preprocessed image into the network to mask image content features, extract high-frequency information, and highlight edge features. Removal and computer-generated detection methods are also utilized.

2.1.1 ELA

Ela can detect whether there is any copy-paste tampering in JPEG image files. By performing two rounds of JPEG compression on an image, if the compression levels of the first and second rounds are similar, there is less information loss, resulting in minimal differences between the two compressed images. Conversely, if there is a significant difference in compression levels, there is more information loss, leading to greater differences between the two compressed images. When an image is tampered with, the compression levels in the tampered regions may differ, or different compression grids may be used. If the tampered image is compressed again using JPEG compression, the differences between the compressed image and the tampered image may vary in the tampered and background regions. This analysis can be used to determine whether the image has been tampered with. When compressing the tampered image again using JPEG compression, if a compression quality factor close to the background region's quality factor is chosen, the differences in the background region will appear darker in the image, while the tampered region may become brighter. The closer the brightness values, the smaller the differences.

2.1.2 EXIF

Exchangeable Image File Format (Exif) is a metadata standard embedded in digital camera photos. It can record various shooting parameters, thumbnail images, and other relevant attributes of the digital camera photos. The implementation of Exif is a file format that does not exist independently. It is attached to multimedia files. For compressed images captured by digital cameras, the Exif format is attached to the APP1 segment of the JPEG format. This segment is one of the application markers of the JPEG format and only records additional information without affecting the decoding and quality of the image itself. Generally, image software tends to ignore the content of this segment, and exif information may be ignored through editing and saving with image software.

### 2.1.3 Color Space

Refers mainly to the YCbCr color space. YCbCr (sometimes also called YUV) is a series of color spaces that represent images and videos. Usually, the signals of the three channels in RGB are strongly correlated, and it is difficult to explain which channel's information is more important. YCbCr is a reversible transformation of RGB signals point by point. The Y signal represents luminance information, and Cb and Cr represent chroma information. Tampering with the YCbCr space may cause changes in brightness, contrast, color, and other aspects of the image, thereby affecting the quality and authenticity of the image.

### 2.1.4 DCT

Discrete Cosine Transform (DCT) is widely used in signal and image processing for lossy data compression. The principle is that when a digital image is compressed using JPEG compression, the DCT parameter quantization process can be regarded as generating a watermark that can detect its integrity. Local tampering of the image will make the DCT values of the tampered area inconsistent with those of the adjacent area. Therefore, comparing the mapped DCT parameters can detect abnormal parts of the DCT values in the image and determine whether there are tampered areas and locate the range of the tampered areas. 2.1.5 Noise Using spatial domain denoising methods such as Gaussian filtering, median filtering, and Wiener adaptive filtering to denoise the original image, the noise map of the image can be obtained by subtracting the denoised image from the original image. If two ROI contents within the image are similar, but the noise maps show inconsistent noise, tampering may exist.

### 2.1.6 Edges

In the same image, objects with the same depth of field should have similar types and degrees of edge blur. Edge detection can determine the degree of blurring of object contour information in the image. If the calculated edge blurring of two objects under different depths of field is different, tampering may exist.

Based on our observations in actual business, using a combination of traditional tampering detection operators for filtering is very useful. We recommend the sherloq toolkit.

### 2.2 Audit Point Localization

The problem of audit point localization is essentially a text detection problem. For different scenarios, the contents of audit point detection are different. Our scenes are mainly divided into tables and documents. For documents, the audit points are usually in specific areas, while for tables, full-image text detection is often chosen. EAST[12] simplifies the pipeline into two steps: the first step is to predict geometric information, foreground/background prediction through a multi-channel FCN; the second step is NMS, which ultimately produces multi-directional text detection results. By modeling the main information of position, scale, direction, and other text through a single network and adding multiple loss functions, that is, the so-called multi-task training, and by predicting geometric information, the results of text detection can be obtained. Although Centernet[13] is not a text detection algorithm, it directly supervises the center point, regresses width, height, and offset, and finally assembles them into bbox. It does not require complex post-processing, but Centernet has center point deviation when detecting long audit points, so some frame connection post-processing needs to be added. In text detection networks based on segmentation, the final binary map is obtained using a fixed threshold, and different thresholds have a significant impact on performance. In DBNet[14], adaptive binarization is performed for each pixel, and the binarization threshold is learned by the network, thoroughly integrating the binarization step into the network for joint training. This makes the final output graph very robust to threshold variations. We ultimately chose DBNet for audit point localization.

2.3 Tampering Detection

2.3.1 Deep Learning Approaches

The current mainstream approach is natural scene tampering detection.

a. Modifying CNN, as CNN is not directly suitable for tampering detection. This is because complex image modification tools can make the modified image visually and statistically similar to the original image. The first layer of CNN is used as a preprocessor, controlling the image content input into the network. The weights are initialized as high-pass filters for calculating residual images in the spatial domain model. This is similar to the rich media mode of image steganography, combining statistical region merging (SRM) with high-pass filters. This method effectively captures sharp edges introduced by tampering operations, especially in splice operations, which are clearly manifested after high-pass filtering. Constrained CNN [15] is used to extract predicted residual features (as the predicted residual largely does not contain image content, but rather tampering traces). Constrained CNN is placed at the beginning of CNN, and higher-level features are formed from these residuals. This suppresses the image content and adaptively learns low-level residual features that are most suitable for detecting forensic evidence. Higher-level forensic features are learned from the residual of deep CNN. RGBN uses noise flow to extract tampering detection features, inputting it into the dual-stream Faster R-CNN for tampering area detection. The difference between tampered and non-tampered areas is captured in the RGB dimension, capturing visual inconsistencies similar to tampering boundaries and the contrast effects between tampered and non-tampered areas, while the noise flow dimension analyzes the noise flow information in the image.

b. Tampering detection can be achieved by utilizing auxiliary frequency domain information or edge information. In [16], the concept of "awareness of frequency" in tampering detection is proposed, as the frequency domain can effectively describe subtle forgery artifacts or compression errors. F3net [17] presents two points: 1. frequency-aware decomposed image components, and 2. local frequency statistics, using DCT for frequency transformation, which has better frequency energy concentration compared to DFT transformation. In mvssnet [18], the key challenge in image tampering detection is how to learn generalized features sensitive to new data tampering while preventing false alarms on real images. Current research emphasizes sensitivity while neglecting specificity, and proposes solutions through multi-view feature learning and multi-scale supervision. This approach is common in tampering detection, similar to the addition of a Fourier transform branch for refined supervision in liveness detection, aiming to distinguish tampered areas on the original image by focusing on tampered image artifacts and edges. Multi-view learning aims to exploit noise distribution and boundary artifacts surrounding tampered regions to learn semantically invariant features for obtaining more universal features, allowing learning from real images, which are not crucial for current semantic segmentation-based methods. ManTraNet is an end-to-end network that performs detection and localization without additional preprocessing and post-processing. It is a fully convolutional network that can handle images of any size and many known forgery types, including splicing, copy-move, deletion, enhancement, and even unknown types. This paper makes two outstanding contributions: it designs a simple and effective self-supervised learning task to classify 385 types of image manipulation for learning robust image manipulation trajectories. Additionally, it formulates the forgery localization problem as a local anomaly detection problem, designs a Z-score feature to capture local anomalies, and proposes a new long short-term memory solution for evaluating local anomalies.

The core idea can be divided into two types: one is to convert it into a classification problem, and the other is to convert it into a segmentation problem, depending on the scene. If only judging whether an image is tampered with, it is sufficient to convert it into a classification problem. However, if you want to know the specific location of the image tampering, segmentation needs to be used, which is itself a pixel-level classification. However, classification detection has a risk that the features of document tampering are relatively weak. Text PS tampering detection includes differences in features such as noise, texture, morphology, and boundary artifacts. But due to the lack of clear positioning indications, the final feature is compressed into a 1-dimensional vector, which may cause the recognition of tampering to be affected by irrelevant tampering features in positive and negative samples.

In addition, many deep learning-based methods create their own tampering detection datasets, which have poor generalization performance and do not work well when applied to different datasets. Moreover, the generation patterns of tampering detection data in natural scenes differ significantly from those in document scenarios.

2.3.2 Document Image Tampering Detection

Document tampering detection has received little research attention and there are almost no available datasets for this task. The tampered regions in document images are small, the image categories vary greatly, and the physical interpretability is weak. There is an extreme data imbalance issue, with a significant class imbalance between normal and abnormal instances, with a ratio that can reach 1000:1. The granularity of tampered regions is very small (<1/100, 16*16 pixels), similar to small object detection in object detection tasks. Additionally, accurately annotating real-world data for tampering detection is challenging, as it is difficult for human observers to identify whether an image has been tampered with and the corresponding regions. Auditors also require auxiliary tools to assist in determining whether tampering has occurred. Furthermore, there are unknown adversarial attacks, where the detection methods naturally lag behind the attacking techniques, which can be highly diverse and unpredictable.

3.Dataset

In document scenarios, some cropping is necessary because document scenes are inherently dense and complex. Directly segmenting the entire image using a natural scene segmentation approach is challenging, especially when areas like monetary values disappear during downsampling. Even with auxiliary prior information added to feature maps, it is difficult to capture tampered regions, similar to the challenge of detecting small objects in traditional object detection tasks. These small objects may not be occluded or obscured but present difficulties in highly similar object recognition. Initially in the project, we discussed single-stage and two-stage approaches. The single-step approach involves integrating tampering localization and recognition into a unified architecture, while the two-step approach first performs tampering detection before recognition. Due to the aforementioned reasons, the two-step approach was ultimately chosen.

3.1 Suning Scene Data

We are working on tampering detection tasks for the supervision, finance, and human resources departments at Suning. The data is broadly categorized into document types and table types. Document types include: account introduction letters, sales supplementary confirmation letters, power of attorney documents, while table types include: price documents, supplier statements, return confirmation forms, return appointments, screenshots of settlement statements, and tractor order templates.

The auditing points for document types vary based on the specific document content. Introduction letters include information such as dates, organizations, account numbers, amounts, names, contact numbers, delivery addresses, ID numbers. Contracts include details like supplier information, recipient information, quantities, unit prices, amounts, names, dates (signing time), addresses, contract numbers, recipients, and email addresses. For table types, we assume comprehensive verification due to the predominantly numerical nature of table documents.

3.2 Dataset Filtering Method

3.2.1 Photos or Scanned Documents

Currently, support is available for scanned documents, while photos require additional processing. The proportion of photos in different types of documents varies. At present, Alibaba categorizes problematic documents as photo-scanned documents, including fake images with real credentials (including fake images with genuine credentials) and real images with fake credentials. The category of fake images with genuine credentials includes tampering detection, tampering localization, and expert rule detection, while the category of real images with fake credentials includes expert rule verification, material background detection, secondary printing detection, and semantic analysis. We have performed binary classification for photos and scanned documents, with scanned letters and paper documents scanned through a scanner showing the most prominent feature of consistent and obvious background noise throughout the entire image.

3.2.2 Electronic

Waybills or PS Documents Currently, support is available for PS documents, while electronic waybills are to be excluded. Electronic waybills are singled out mainly because they lack obvious background noise in digital files, making it difficult to extract differences due to the absence of noise contrast throughout the entire image, resulting in poor noise effect.

3.3 Attack Methods

Scaling: Generally, the greater the compression ratio, the lower the image quality, making it appear less clear and causing color distortion. It can be said that there is an inverse relationship between image quality and compression ratio. Enhancement: Such as contrast and brightness adjustments. Social transmission: Transmission via platforms like WeChat results in significant distortion from the original image. Anti-forensics: Targeted noise and trace elimination. Adversarial examples: Targeting network and feature attacks. However, in document tampering, the dominant and core method is still PS tampering detection. The broad category of PS includes operations using software such as Photoshop, Meitu XiuXiu, including same-image tampering and splicing and erasure operations. For example, stamps in introduction letters in the process are almost identical, and these are extracted using PS.

3.4 Data Augmentation

a. Enrich pre-processing enhancements that do not change semantics, such as adjusting brightness, contrast, and optical distortion.

b. Automatic generation tool for tampering with text areas, including text splicing, deletion, addition, replacement, copy-pasting, etc.

c. Change a portion of the amount in a different way, while keeping the actual amount unchanged.

d. Erase while modifying, leaving traces at the bottom.

e. Direct erasure.

f. Obvious tampering in Photoshop.

g. Cut a portion and paste it onto another, splicing.

h. Uniformly crop, apply Gaussian blur, change background information, then paste back, which means finding the target area, cropping, and pasting back.

i. Stretch single characters, 0.8/0.9/1/1.1/1.2/…,

However, the actual complexity of tampering in business scenarios is much higher than that in the training set as a whole.

4. Proposed Model

This is an overall design and solution for this type of problem in Suning scenarios. The cascade solution first filters out some documents that do not meet the requirements, such as brightness and size requirements.

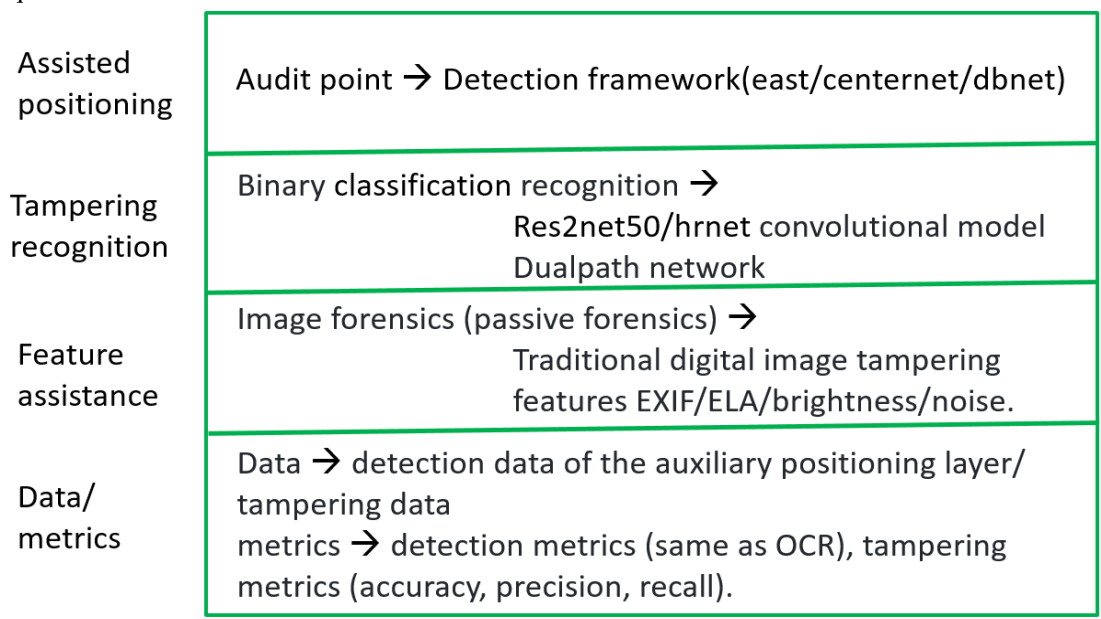

Fig2: Proposed method's tampering detection framework diagram.

The process diagram in Suning's business scenario is shown below.

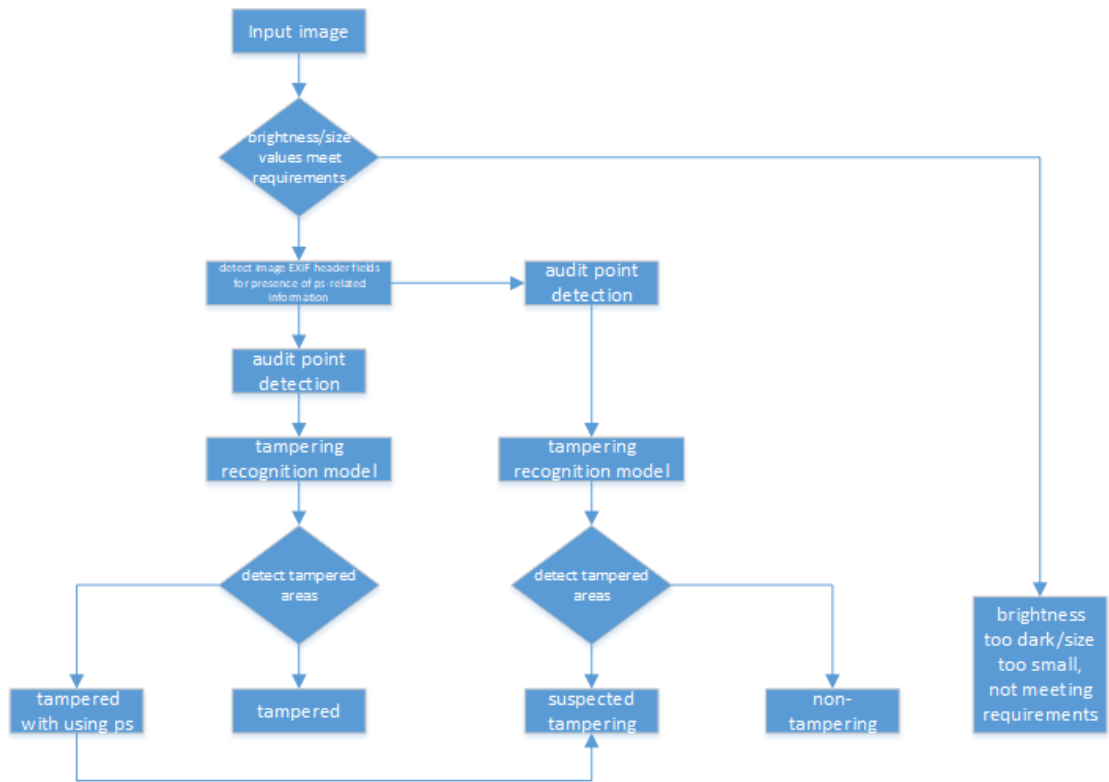

Fig 3: Business scenario process diagram

4.1 Feature-assisted

The advantage of the feature-assisted layer lies in the fact that the failure of the two-stage algorithm does not necessarily indicate a misjudgment of tampering, because the image itself has still been modified using software such as Photoshop, but the tampered area may not necessarily be at the amount (it could be the date, seal), and even some quotations made in Photoshop were detected in non-standard data.

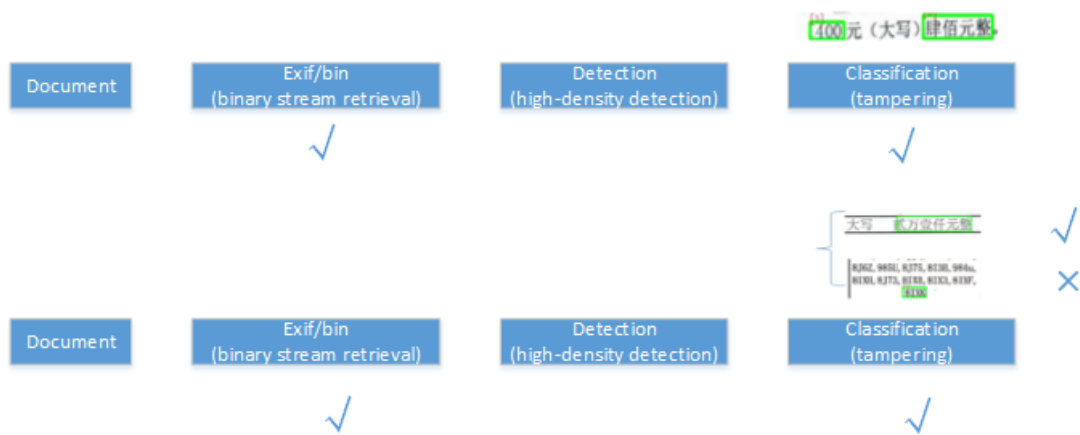

Fig 4: The first line represents standard data, indicating the completion of the entire detection process. The second line represents non-standard data, meaning that the first-stage and second-stage models of tampering may generate a certain degree of false detection due to the mismatch in training data.

4.2 Audit Point Localization

We employed improved versions of EAST, CenterNet, and DBNet for different business scenarios. Among them, CenterNet supports detection of inclined bounding boxes. Ultimately, we chose the DBNet model for text detection. For documents, we detect the corresponding audit point fields, while for tables, we detect all fields.

4.3 Tampering Recognition
4.3.1 Data cropping method

Cropping strategies: V0 - Center crop, V1 - Sequential crop, V2 - Right-side crop, V3 - Direct resize. In our algorithm, we chose crop 1.

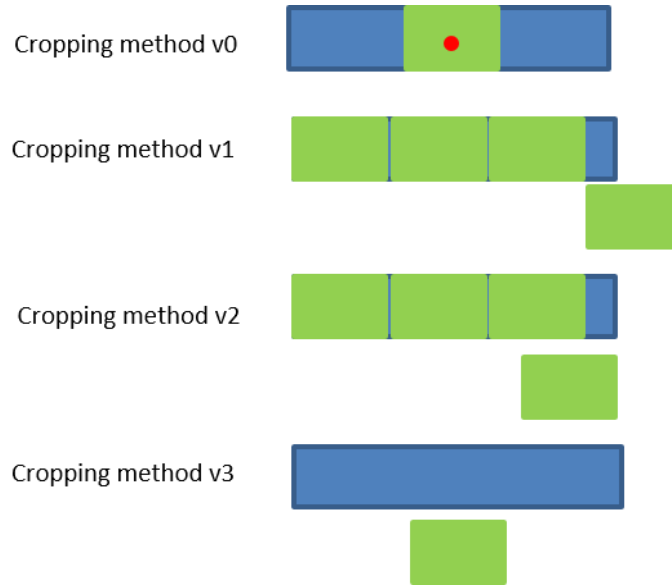

Fig 5: Illustration of cropping methods.

4.3.2 Dual-path Tampering Recognition network

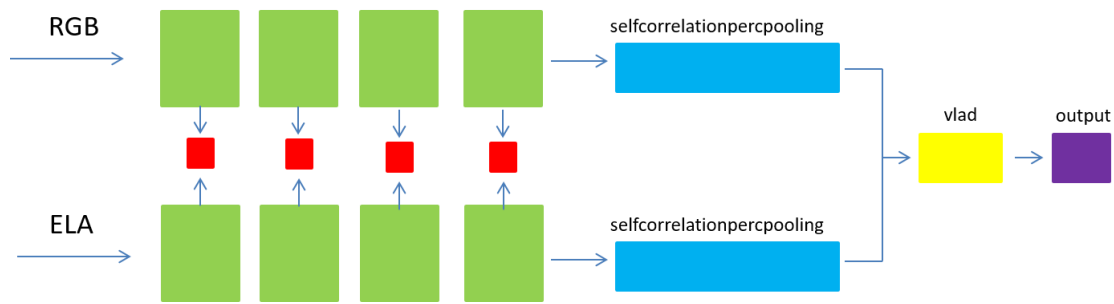

Fig 6: Illustration of the dual-stream tampering detection network

a. Dual-path attention Res2Net

For the original RGB image and the image processed with ELA, we designed a dual-path network structure to collaboratively process input information from two different domains. The overall network structure is shown in the figure above, where ResNet18 is used to handle RGB and ELA images separately. In the middle of the model, we introduced SE modules to apply attention mechanism for feature enhancement and to fuse features between the two models. Finally, the features output by the two models are reduced and clustered using NetVLAD, and the concatenated features are then used for classification in the final FC layer.

b. NetVLAD

VLAD (Vector of Locally Aggregated Descriptors) is a feature representation method for images that is similar to Bag-of-Features (BOF). It can be understood as an encoding method that represents local features as global features. This layer can be easily embedded into any CNN architecture and can be trained through backpropagation. The VLAD structure can re-encode the features extracted by a convolutional model, and features encoded by VLAD have better global expressive power, which benefits the model in judging tampered images.

NetVLAD layer can be directly connected to the last layer of the convolutional network (H×W×D), treating the feature map of the last layer as N dense D-dimensional local descriptors. NetVLAD replaces the original Global Average Pooling and brings about certain improvements in the final classification performance.

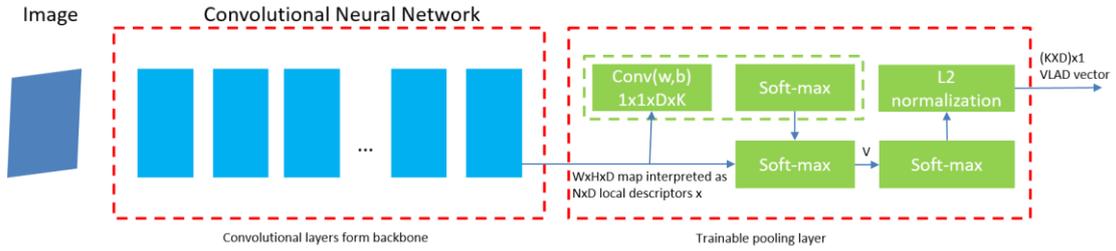

Fig 7 Illustration of the NetVLAD network structure.

c. Similarity Detection Block

One of the drawbacks of the above-mentioned ELA technology is its insensitivity to self-replicating modifications in images. To address this issue, a Similarity Detection Block [19] is introduced. This block utilizes the features extracted by CNN. The Self-Correlation layer calculates the similarity between features, and then the Percentile Pooling extracts features with local representativeness, resulting in more descriptive features.

5.Experiments

5.1 Auxiliary Point Localization

We attempted various detection methods, and the evaluation metrics for the detection are as follows. We used two approaches, TIOU [20] and DetEva. In the table below, we included centernet, PMTD [21], BDN [22], and dbnet. The "data" column represents letter data, while "forms" represents form data. We separately trained the models on form and letter scenarios, but ultimately chose to incorporate them into a single detection model, which yielded satisfactory results.

Tab 1: Accuracy Table of Auxiliary Point Localization Model

| model | data | backbone | conf | TIOU | | | DetEva | | |
|---|---|---|---|---|---|---|---|---|---|
| | | | | F1-score | Precision | recall | F1-score | precison | Recall |
| centernet | letters | Resnet50 | 0.1 | 0.5475 | 0.6834 | 0.4567 | 0.6133 | 0.7277 | 0.53 |
| PMTD | letters | Resnet50 | 0.5 | 0.4242 | 0.4895 | 0.3702 | 0.4082 | 0.4401 | 0.3805 |
| BDN | letters | R-50-FPN | 0.9 | 0.6976 | 0.7591 | 0.6453 | 0.7711 | 0.7897 | 0.7534 |
| dbnet | all data | resnet18 | 0.2, 0.45 | 0.7198 | 0.7386 | 0.7019 | 0.8402 | 0.8309 | 0.8497 |
| | form | | | 0.7742 | 0.8046 | 0.7461 | 0.8842 | 0.8935 | 0.8751 |
| | letters | | | 0.6377 | 0.6441 | 0.6315 | 0.7738 | 0.7412 | 0.8093 |

5.2 Tampering Recognition model

We used the dual-path recognition network. Table 2 shows the model configuration, while Table 3 displays the corresponding accuracy results. It can be seen that our tampering detection model has good accuracy and recall.

Tab 2: Parameters and Model Configuration Table for Tampering Recognition Model

| model | Lr | origin | ela | nosie | feather vector | fusion | Dim reduction |
|---|---|---|---|---|---|---|---|
| dpv1 | 0.00035 | resnet18 | resnet18 | | avgpool | add | |
| dpv2 | 0.00035 | resnet18 | resnet18 | | selfcorrelationpercpooling | concat | vlad |
| dpv2.1 | 3e-4 | resnet18 | resnet18 | | selfcorrelationpercpooling | concat | vlad |
| dpv2.2 | 0.00035 | res2net50 | resnet18 | | selfcorrelationpercpooling | concat | vlad |
| dpv2.3 | 3e-4 | res2net50 | res2net50 | | selfcorrelationpercpooling | concat | vlad |
| dpv3 | 0.00035 | resnet18 | resnet18 | srm | selfcorrelationpercpooling | concat | vlad |

Tab 3: Accuracy Table for Tampering Recognition Model

| model | accuracy | recall | precision | tp/fp/fn/tn | time(s) |
|---|---|---|---|---|---|
| dpv1 | 0.61 | 0.338 | 0.869 | 93/14/182/215 | 0.0580 |
| dpv2 | 0.8909 | 0.964 | 0.855 | 265/45/10/184 | 0.0688 |
| dpv2.1 | 0.9164 | 0.863 | 0.875 | 251/36/40/582 | 0.0725 |
| dpv2.2 | 0.7421 | 0.789 | 0.751 | 217/72/58/157 | 0.07 |
| dpv2.3 | 0.8016 | 0.913 | 0.768 | 251/76/24/153 | 0.0826 |
| dpv3 | 0.7361 | 0.676 | 0.809 | 186/44/89/185 | 0.061 |

As we trained and tested in the document tampering detection scenario, we also conducted training and testing on CASIA2 to evaluate the generalization of the detection methods.

| model | accuracy | recall | precision | tp/fp/fn/tn | time(s) |
|---|---|---|---|---|---|
| dpv1 | 0.8974 | 0.93 | 0.836 | 963/187/72/1312 | 0.11 |
| dpv2 | 0.8657 | 0.862 | 0.817 | 884/198/141/1301 | 0.09533 |
| dpv3 | 0.8974 | 0.93 | 0.836 | 953/187/72/1499 | 0.139 |

Examples of Partial Tampering Detection. Due to the involvement of sensitive information such as brands and merchants, only some tampered data on the amounts are displayed.

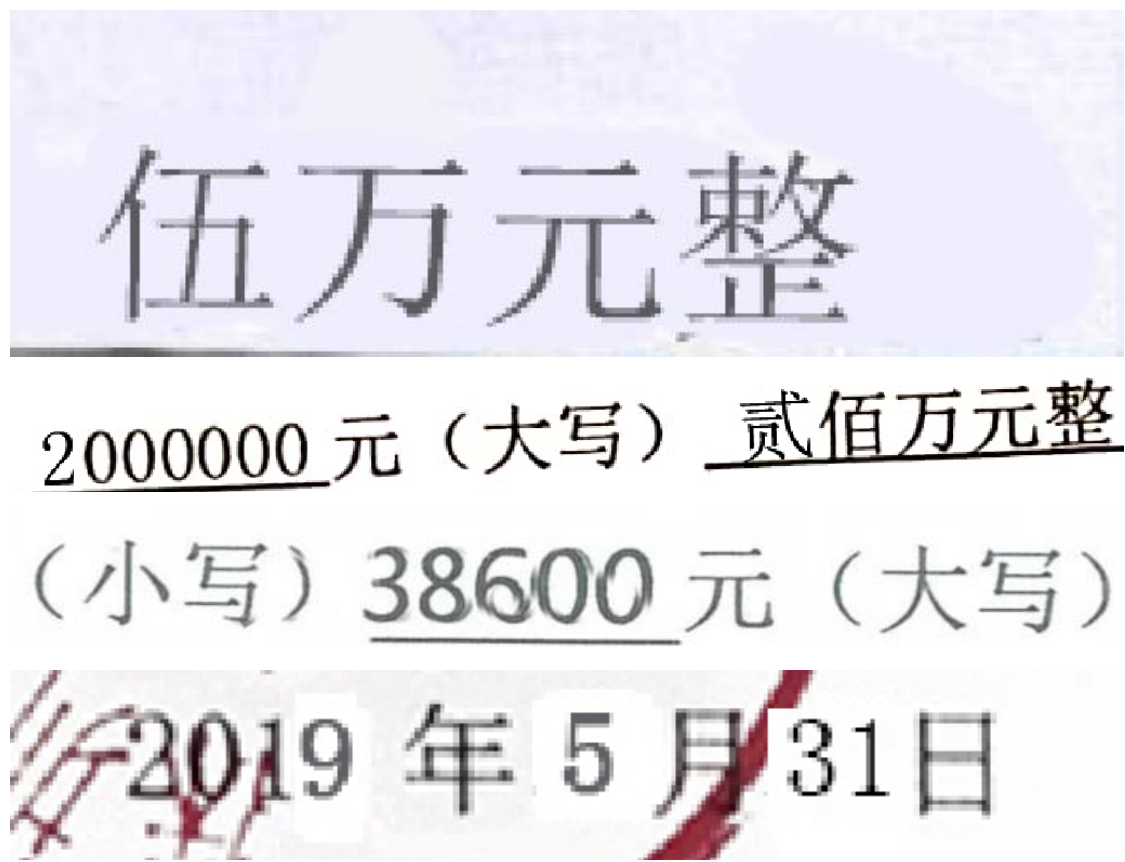

Fig 8: Tampering Example Image

6.conclusion

  Detecting document tampering has always been one of the most challenging aspects of tampering detection. The two-stage dual-path Ps tampering detection and recognition method we propose includes three steps: feature assistance, audit point localization, and tampering recognition. It involves hierarchical filtering and graded outputs (tampered/suspected tampered/untampered). Feature assistance includes exif/binary stream keyword retrieval/noise, with branch detection based on the results. Audit point localization utilizes detection frameworks to control thresholds for high and low-density detection. Tampering recognition involves a dual-path recognition network, extracting features from RGB and ELA streams, reducing dimensions through self-correlation percpooling, and outputting after fusion via vlad. Significant achievements have been made in various fields at Suning, including supervision, finance, and personnel, greatly reducing occurrences of document tampering incidents.